\documentclass[conference]{IEEEtran}
\usepackage{cite}
\usepackage{amsmath,amssymb,amsfonts}
\usepackage{algorithmic}
\usepackage{graphicx}
\usepackage{textcomp}
\usepackage{multirow}
\title{ Graph-based Question Answering System }
\def\BibTeX{{\rm B\kern-.05em{\sc i\kern-.025em b}\kern-.08em
    T\kern-.1667em\lower.7ex\hbox{E}\kern-.125emX}}
    
\makeatletter
\renewcommand{\@IEEEsectpunct}{\ \,}
\makeatother
\begin{document}

\author{\IEEEauthorblockN{Piyush Mital}
\IEEEauthorblockA{\textit{College of Engineering, Pune} \\
\textit{mitalpk14.comp@coep.ac.in}
}
\and
\IEEEauthorblockN{Saurabh Agrawal}
\IEEEauthorblockA{\textit{College of Engineering, Pune} \\
\textit{agrawalss14.comp@coep.ac.in}
}
\and
\IEEEauthorblockN{\hspace{0cm}Bhargavi Neti}
\IEEEauthorblockA{\hspace{0cm}\textit{College of Engineering, Pune} \\
\textit{\hspace{0cm}bhargavinv14.comp@coep.ac.in}}
\and
\IEEEauthorblockN{\hspace{40pt}Yashodhara Haribhakta}

\IEEEauthorblockA{\hspace{40pt}\textit{College of Engineering, Pune} \\
\textit{\hspace{40pt}ybl.comp@coep.ac.in}}
\and
\IEEEauthorblockN{\hspace{10pt}Vibhavari Kamble}
\IEEEauthorblockA{\hspace{10pt}\textit{College of Engineering, Pune} \\
\textit{\hspace{10pt}vvk.comp@coep.ac.in}
}
\and
\IEEEauthorblockN{\hspace{0pt}Krishnanjan Bhattacharjee}
\IEEEauthorblockA{\hspace{0pt}\textit{C-DAC, Pune, India} \\
\textit{\hspace{0pt}krishnanjanb@cdac.in}
}
\and
\IEEEauthorblockN{\hspace{60pt}Debashri Das}
\IEEEauthorblockA{\hspace{60pt}\textit{C-DAC, Pune, India} \\
\hspace{60pt}\textit{debarshid@cdac.in}
}
\and
\IEEEauthorblockN{\hspace{70pt}Swati Mehta}
\IEEEauthorblockA{\hspace{70pt}\textit{C-DAC, Pune, India} \\
\hspace{70pt}\textit{swatim@cdac.in}
}
\and
\IEEEauthorblockN{\hspace{70pt}Ajai Kumar}
\IEEEauthorblockA{\hspace{70pt}\textit{C-DAC, Pune, India} \\
\textit{\hspace{70pt}ajai@cdac.in}
}

  }




    \maketitle
    \begin{abstract}
     \textbf{ In today's digital age, in the dawning era of big data analytics, it is not the information but the linking of information through entities and actions, which defines the discourse. Any textual data either available on the Internet or off-line (like newspaper data, Wikipedia dump, etc) is basically connected information which cannot be treated isolated for its wholesome semantics.There is a need for an automated retrieval process with proper information extraction to structure the data for relevant and fast text analytics. The first big challenge is the conversion of unstructured textual data to structured data. Unlike other databases, graph databases handle relationships and connections very elegantly. Our project aims at developing a graph based information extraction and retrieval system.\\\\}
    \end{abstract}

\section{Introduction}
Information retrieval becomes easier when the unstructured data is represented in the form of a graph with entities as nodes and edges representing their semantically relevant connections. They are also being used by applications like Google Now, Microsoft Cortana and Apple Siri which are capable of understanding natural language queries and answer questions, making recommendations, etc. to the user. Graph development is thus a major step towards effecting intelligent machines. {\\}
In order to develop such a robust question and answer system, we need to create a thorough graph by accurately resolving all relationships and converting the given question into a structured query, improving our understanding of the question.  For this, we need to hone relation extracting and co-referencing tools and incorporate a series of text and linguistic based techniques on our graph database system including tokenisation, named entity recognition, parts of speech tagging, semantic role assignment  extractor, WordNet normalization etc.

\section{System Architecture}

The project involves two main phases - {\\}
Firstly, XML and English text to graph database creation and secondly, querying and visualization. In the first phase, we build a graph of entities and relationships from given input XML and unstructured English sentences. We have identified patterns in sentences to extract relationships between entities. {\\}
In the second phase, we query the graph database with natural language queries to retrieve results. The queries are annotated using different database oriented classifications to be processed for graph creation. The query graph is matched with the document graph, created in the first phase and the matched results are visualized with a graphical depiction of entities and their relations so as to give better and quicker insight into the information. {\\}

\subsection{QUERY PROCESSING}

\begin{figure}[htbp]
\centerline{\includegraphics[width=7cm,height=7cm]{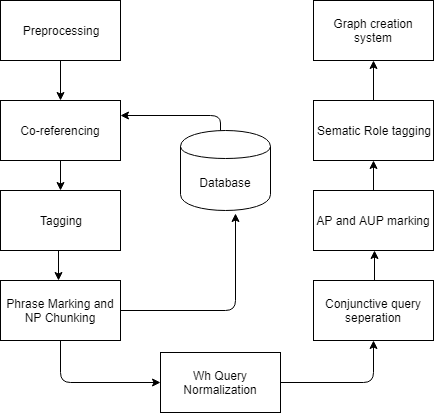}}
\caption{The query processing flow diagram}
\label{fig}
\end{figure}
\subsubsection{ \textbf{Preprocessing\\} } 

\begin{itemize}
\item Dot Handling : Dot Handling is done for all abbreviations like Mr., Dr., Rs. etc. 

\item Garbage Removal : Within this step, unnecessary words/ phrases/symbols (whose value is equivalent to garbage) are removed.
\begin{itemize}
\item Special Character removal : 
Special characters like [*], [?], [@] are removed in this step ([,], [-], [‘] are retained ). 

\item Remove ignorable phrases :
Ignorable patterns like “give me details “, “give me report” etc. in a query, whose meaning is implicit, are removed from the query string.
\end{itemize}

\item Date-Time Normalization :
The same date-time normalizer that was used while document processing is used on the query, so that there is no difference in tagging and conventions.
\end{itemize}

\subsubsection{\textbf{Co-referencing\\}}
The query is passed through the Spacy co-referencing tool, where all the pronouns in the query are replaced with the noun they are referring to. 

\subsubsection{\textbf{Tagging}}

\begin{itemize}

\item POS and NER Tagging :
The output of Step 2 is passed through Stanford CoreNLP, and  POS and NER tagging of each word is done. 

\item Database(Db)-tagging :
The query is tagged with the phrases and NER tags present in the database - namely names(NEP tag),  Countries (NEC tag), Locations (NEL tag), Prepositions(xprep,aprep,etc tags), Concept words(S, K, Y tags), Common Nouns (CN tag), Political Designations(NED tag), Political Parties(NEPT tag) and Organisations(NEO tag). 
\end{itemize}
\subsubsection{\textbf{Phrase Marking {\\}} } 
The algorithm for phrase marking is as follows :
\begin{itemize}
\item 3  phrase lists are created : phrases extracted from POS tags, phrases extracted from NER tags and phrases matching with those present in the database.
\begin{itemize}
\item Group words having POS tag NNP consecutively into a phrase
\item Group words having the same NER tag consecutively into a phrase
\item Match words from tables in the database with words in the query
\end{itemize}
\end{itemize}
\begin{itemize}
\item Union of these 3 phrase lists is done and a final list of phrases having both POS and NER tags is made
\begin{itemize}
\item If a phrase is repeated in the NER phrases and Db phrases, then preference is given to the tag in Db phrase list
\end{itemize}
\item The query is re-created by embedding these phrases into the preprocessed and then inserting their POS and NER tags
\end{itemize}

\subsubsection{\textbf{Noun Phrase (NP)-Chunk Creation {\\}} }

NP-Chunks are created from sequence of noun words, with a combination of a sequence of adjective and adverbs. Chunks are either loosely or strictly bound. This is done to give more added descriptive properties to the named entities. Once the NP-Chunks are formed, they, along with the phrase marked chunks, form the Named Entities of the system.

\subsubsection{\textbf{'Wh-query' to normal query\\}}

 If any '/WP' POS tag is found in the query string which is present at the beginning of the query  then it is treated as a 'Wh-query' i.e. a query that asks a Who, What, When or Where question and is sent for processing. If, in between the sentence '/WP' occurs, then it is not a wh-query.

\subsubsection{\textbf{Conjunctive Query Separation\\}}

Either input from Co-referencing module after named entity replacement in place of Pronouns like, "Ram's visit and his statement - Ram's visit and Ram's statement" or from natural multiple concept query, like "statement and meeting of Sita" or multiple sequence of named entity query like "statement of Ram and Sita" as well as multiple concept query like "statement and visit of Ram and Sita" are resolved in this step. 

\subsubsection{\textbf{Agent Phrase (AP) and Acted Upon Phrase (AUP) Marking\\ }}
These are Noun phrases and Verb phrases present in the sentence which can be replaced by Dummy entities to facilitate the Semantic Role assignment of the Query.{\\}
Example:  Statement of Ram on Shyam’s killing by Ravi {\\}
Main Sentence: Statement of Ram on DUMMY{\\}
DUMMY:Shyam’s killing by Ram

\subsubsection {\textbf{Semantic Roles Application on each Unit Query as applicable\\ }}
 (Agent)A, (Acted upon)AU, (Acted upon Place Destination)AUPlaceD, (Acted upon Place Source)AUPlaceS, (Agent Phrase)AP, (Acted upon Phrase)AUP are to be tagged as per patterns of Semantic role application using Regex table provided by CDAC. Actor Phrases and Acted Upon Phrases (AP/AUP) are also tagged with Semantic roles in this step done. 

  \subsubsection{\textbf{Creation of Graph tree \\}}
After the assignment of semantic roles, the semantic role tagged elements and concepts are fetched from the processed input query along with descriptive relation and entity attributes and given input to graph creation system. The graph of the queries then is visualized in neo4j.

\subsection{DOCUMENT PROCESSING}

\begin{figure}[htbp]
\centerline{\includegraphics[width=7cm,height=6cm]{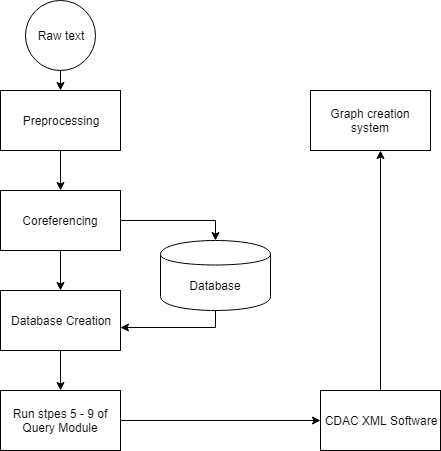}}
\caption{The document processing flow diagram}
\label{fig}
\end{figure} 

\subsubsection {\textbf{Preprocessing\\}}
Steps 1-4 from the query processing module are run. 

\subsubsection {\textbf{Co-referencing\\}}

Using Spacy, all of the documents are co-reference. Manual coreferencing is then performed on the document to provide further accuracy while generating graphs. 

\subsubsection{\textbf{Database Creation}}
\begin{itemize}
\item Tagging :
These documents are passed through Stanford CoreNLP for POS and NER tagging. 

\item Phrase Marking : 
Words with the same POS or NER tag occurring consecutively are grouped together into a phrase. These phrase lists are then stored into the database. 

\item Concept Tagging: 
A specific set of words, mainly verbs and nouns, in the documents are extracted as concept words and then they are classified manually into the S, K and Y action classes. These words along with their action classes are stored into the database. 

\item Collecting database information :
Also, the following tables are also created and stored into the database so as to assist in the query preprocessing steps : Ambiguous Words, Ignorable Patterns, Prepositions, Political Designation, Party Table and Countries.
\end{itemize}
\subsubsection{\textbf{Semantic Role Application\\}}
Pass the document available through CDAC provided software for processing. Repeat steps 5 - 9 of the query module for the document sentences. Following steps are an addition to the query module steps to assign semantic roles to the tagged document sentences.

\subsubsection{\textbf{XML Creation \\}}
Pass this Annotated Document through CDAC software to generate XML output of the text. Additional Tags apart from the ones mentioned in the Query Module introduced in the XML are - 
A - Verb, P - Verb, Copula, SubAction and SubAction\_NETYPE.
\begin{itemize}
\item A - verb and P - verb are categories of main verbs that are considered as graph relations
\item SubActions and SubAction\_NETYPES encompass the actions and entities of preposition phrases of a given sentence.
\item The Noun Phrases and Verb Phrases are Catered in Step 7 of Query Module
\end{itemize}

All of these are Tags are used to create the Graph of the Document.

\subsection{GRAPH CREATION SYSTEM}

Graph of Documents and Queries can be created as follows: 

\hspace{2mm} Relation for Document Graphs - A-verb, P -verb, Copula  

\hspace{2mm} Relation for Query Graphs - (S, K, Y, N) Concept Words 

\begin{itemize}
\item Agent(A) -- [ Relation ]--$>$ Patient(AU)
\item Agent(A) -- [ Relation ]--$>$ Acted Upon Phrase (AUP) 
\item Patient(A) -- [ SubAction ]--$>$ SubAction\_NETYPE 
\item Agent Phrase(AP) -- [ Relation ]--$>$ Patient (AU)
\item Agent Phrase(AP) -- [ Relation ]--$>$ Acted Upon Phrase (AUP)  {\\}
\end{itemize}  
Patient may Consist of a variety of entities - 
AU (Acted Upon), Destination Location (AUPlaceD), Source Location( AUPlaceS), Time, Date, etc

For Agents and Patients with Possessive " 's "  elements, Each Element is tokenized to build a chain of nodes connected by "has a" hierarchy. {\\}

\textbf{Example} : John's Brother killed Ram, becomes {\\}
( John )--[ has a ]--$>$( Brother )--[ killed ]--$>$( Ram ){\\}

All the Relations between any 2 nodes contains the Sentence number and the Document number to which that Node belongs. This enables efficient retrieval of actual sentences that have been represented by graphs.

\subsection{MATCHING MODULE}

\begin{figure}[htbp]
\centerline{\includegraphics[width=5cm,height=5cm]{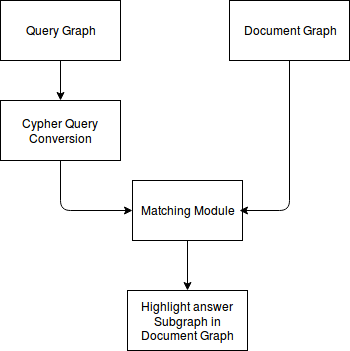}}
\caption{The Graph Matching and Visualization flow diagram}
\label{fig}
\end{figure}

Query graphs created above are converted into Cypher Queries and Run through the Document graph stored in the Neo4j Database.
Matching is facilitated for  Questions like Who, Where, Which Place, What time, Which day / month / year / date,  When and Whom , How many, How Much. 
The motive of the matching algorithm is to provide pin point answers to questions. {\\}

Following are the ways to match and retrieve answers for Wh Queries:
\begin{itemize}
\item \textbf{Who} - Seek an Agent or a Patient with NE Type as Person  (NEP) 
\item \textbf{When}, Which day / month / year / date  - Seek  patient with NE Type as Date and disambiguate 
\item \textbf{What time} - Seek patient with NE Type 
\item \textbf{Where}, Which place - Seek patient with NE Type as Location or Organization
\item \textbf{Whom} - Seeks only Patient or Agent with NE TYPE as Person or Organization
\item \textbf{How many}, How much - Seeks patient with NE Type as NUMEX (Quantity/Money/Percentage)
\item Questions like Give me report on, Show graph on, etc - Match input query directly with the Document graph.
\end{itemize}

\subsection{VISUALIZATION}
D3.js and neo4j database are used to visualize the answers matched in the query module. 
The matched query is highlighted in the Document Graph. Pinpoint answers are retrieved. 

\section{Implementation}
1000 Documents from trending topics like Actor, Singer profiles, Films, Festivals, Sportspersons, Tourist places and Political events are extracted from wikipedia text.
These were passed through the Document Processing module. Document graph was generated out of these documents.
A set of questions of types who, when, where, whom, how much and how many were derived out of the documents' contents and were passed through the query processing module. The cypher query constructed out of the query graph was then run upon the document graph and results were noted down. {\\}

\begin{figure}[htbp]
\centerline{\includegraphics[width=10cm,height=6cm]{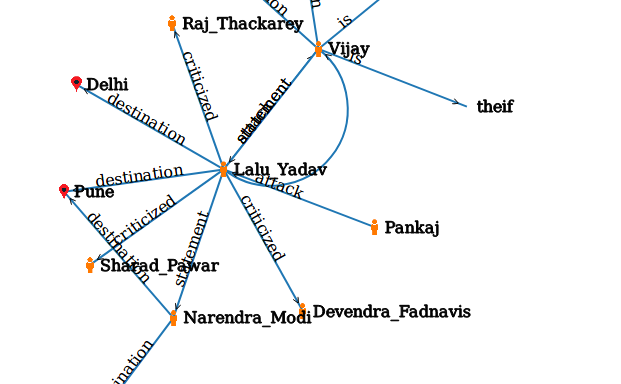}}
\caption{Document Graph before matching}
\label{fig}
\end{figure}

\begin{figure}[htbp]
\centerline{\includegraphics[width=6cm,height=4cm]{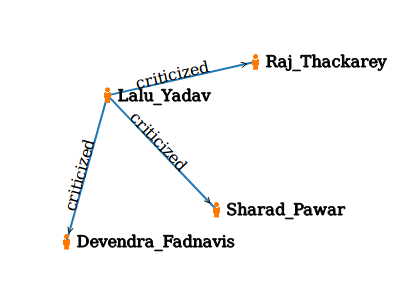}}
\caption{Document Graph after matching}
\label{fig}
\end{figure}

\textbf{Example:} 
In Fig 5. matching was achieved for the query :{\\}
"Who was criticized by Lalu Yadav?"
{\\}Multiple patients were fetched in the matching and have been highlighted in the output graph.

\section{Results}
Results of the system have been recorded in Table 1. {\\}
\begin{itemize}
\item Who, where, when and whom are wh questions derived from the documents. They are used to test the system's accuracy of assigning semantic roles to entities present in the query.
\item Agent , Patient , Source and Destination Resolution are the tests done to adjudge the accuracy of annotating entities with the given semantic role. 
\item Embedded Queries and Conjunction queries  testing are done to adjudge the accuracy of separation of the original query into sub queries as per requirement. 
\end{itemize}
The system has an average accuracy of 80.1 \% (of understanding the questions).
\begin{itemize}
\item The accuracy for "Who" is the best as the cases to be handled is the least.
\item The accuracy of "Patient resolution" is the least as the patient is substituted by date, source location or destination location as per the query
\end{itemize}


\begin{table}[]
\centering
\caption{Results}
\begin{tabular}{llllll}
Query type            & Total queries & Precision & Recall & Accuracy & F-score \\
Who                   & 263           & 88.76     & 89.27  & 85.17    & 89.01  \\
Where / Which place   & 134           & 87.91     & 84.21  & 80.6     & 86.02  \\
When / What time      & 193           & 85.38     & 84.09  & 79.27    & 84.73  \\
Whom                  & 79            & 81.36     & 90.57  & 79.75    & 85.71  \\
Agent Resolution      & 304           & 87.19     & 88.66  & 80.92    & 87.92  \\
Patient Resolution    & 269           & 75.77     & 89.63  & 76.21    & 82.12  \\
Source Resolution     & 295           & 81.5      & 85.34  & 77.97    & 83.33  \\
Destination Resolution& 217           & 90.13     & 80.59  & 79.75    & 85.09  \\
Embedded Queries      & 141           & 90.72     & 82.24  & 80.14    & 86.27  \\
with conjunction      & 295           & 85.78     & 88.21  & 81.02    & 86.98  \\
                      & Average       & 85.45     & 86.28  & 80.1     & 85.71  \\
\end{tabular}
\end{table}

\section{Limitations}
The system has certain limitations which have been described below:
\begin{itemize}
\item It assumes that the English sentences taken as input have a correct grammar with use of commas and conjunctions appropriately.
\item The system does not cater to an alternate form of questions, for eg: the alternate form of “What is the height of Mount Everest?” is “How tall is Mount Everest?”, nor complex sentence structures in the document for eg, those which contain the word 'respectively'. 
\item The system is prone to errors as document processing is dependent on a large extent on the input received by it in the form XML documents and coreferencing, which are not completely correct. 
\item Along with true positives, the system retrieves lots of false positives as well i.e. it tries to answer every question and provides a wrong answer to a question that was wrong in the first place.
\item Some lesser important descriptive elements of the sentence are eliminated during graph creation and only important elements are retained. Therefore the system can cater questions to only those.
\item Graph databases, in general, have a larger space complexity as compared to normal databases. 
\end{itemize}

\section{Conclusion}

We have developed a graph-based information extraction and retrieval system from unstructured natural language text documents to structured graphs along with natural language querying. We developed and infused our classical NLP based text and  linguistic techniques for query processing,  phrase chunking, semantic role extraction, concept extraction etc to better understand our input and generate elements that constitute the graph. We have achieved a fairly good accuracy for a large no of sentence cases. Despite the limitations of the system and graphs in general, the system is a proof of concept to demonstrate the capabilities of graph-databases in question answering systems - Graphs can simply but adroitly represent relationships between different parts of a sentence. They have huge potential in the development of intelligent information systems and can be extended to many future applications. 

\section{Future Work}

\begin{itemize}
\item The System was designed to cater to facts based data/corpus. Defence Ministry of a country has data in the form of facts. Facts contain Person names, Places, Dates, Organizations, and relations between them. Each of these facts is linked to other ones as well. The system can convert all the corpus into the document graph and analyze the relationships between 2 named entities(nodes). 
\item Question validation system can also be created to verify whether the question to a question that had a valid answer or is it a made-up question.  
\item this application can be extended to document(s) summarization, since the NLP based query processing and document processing in this question answering system helps us retrieve answers to key questions in the content, by analyzing and understanding them in great depth, and could address causal questions ("Why" and "How"). 
\end{itemize}

\section*{Acknowledgment}

We would like to thank our internal guides, Mrs. Vibhavari Kamble and Dr. Yashodhara Haribhakta for their constant guidance, evaluation and support for this project. We would also like to thank our external guides from CDAC, Dr. Krishnanjan Bhattacharjee, Dr. Ajai Kumar, Ms. Swati Mehta, Mr. Debarshi Das for giving us direction and feedback.

\end{document}